\documentclass[10pt,twocolumn,letterpaper]{article}

\newcommand{\Tref}[1]{Table~\ref{#1}}

\newcommand{\Fref}[1]{Figure~\ref{#1}}
\newcommand{\Sref}[1]{Section~\ref{#1}}

\usepackage{authblk}

\usepackage[pagenumbers]{cvpr} 

\usepackage{graphicx}
\usepackage{amsmath}
\usepackage{amssymb}
\usepackage{booktabs}
\usepackage{color}
\usepackage{multirow}
\usepackage{verbatim}
\usepackage{indentfirst}

\usepackage[pagebackref,breaklinks,colorlinks]{hyperref}

\usepackage[capitalize]{cleveref}
\crefname{section}{Sec.}{Secs.}
\Crefname{section}{Section}{Sections}
\Crefname{table}{Table}{Tables}
\crefname{table}{Tab.}{Tabs.}

\makeatletter
\newcommand{\printfnsymbol}[1]{%
  \textsuperscript{\@fnsymbol{#1}}%
}
\makeatother

\def\cvprPaperID{5856} 
\def\confName{CVPR}
\def\confYear{2022}

\begin{document}

\title{UBoCo : Unsupervised Boundary Contrastive Learning\\for  Generic Event Boundary Detection}
\author{Hyolim Kang \thanks{equal contribution, ordered by surname}}
\newcommand\CoAuthormark{\footnotemark[\arabic{footnote}]}
\author{Jinwoo Kim \protect\CoAuthormark}
\author{Taehyun Kim}
\author{Seon Joo Kim}
\affil{Yonsei University}
\affil{\small{\texttt{\{hyolimkang,jinwoo-kim,kimth0101,seonjookim\}@yonsei.ac.kr}}}
\maketitle

\begin{abstract}
Generic Event Boundary Detection (GEBD) is a newly suggested video understanding task that aims to find one level deeper semantic boundaries of events.
Bridging the gap between natural human perception and video understanding, it has various potential applications, including interpretable and semantically valid video parsing.
Still at an early development stage, existing GEBD solvers are simple extensions of relevant video understanding tasks, disregarding GEBD's distinctive characteristics.
In this paper, we propose a novel framework for unsupervised/supervised GEBD, by using the Temporal Self-similarity Matrix (TSM) as the video representation.
The new Recursive TSM Parsing (RTP) algorithm exploits local diagonal patterns in TSM to detect boundaries, and it is combined with the Boundary Contrastive (BoCo) loss to train our encoder to generate more informative TSMs. 
Our framework can be applied to both unsupervised and supervised settings, with both achieving state-of-the-art performance by a huge margin in GEBD benchmark. 
Especially, our unsupervised method outperforms the previous state-of-the-art ``supervised'' model, implying its exceptional efficacy.
\end{abstract}


\begin{figure*}
  \includegraphics[width=\textwidth]{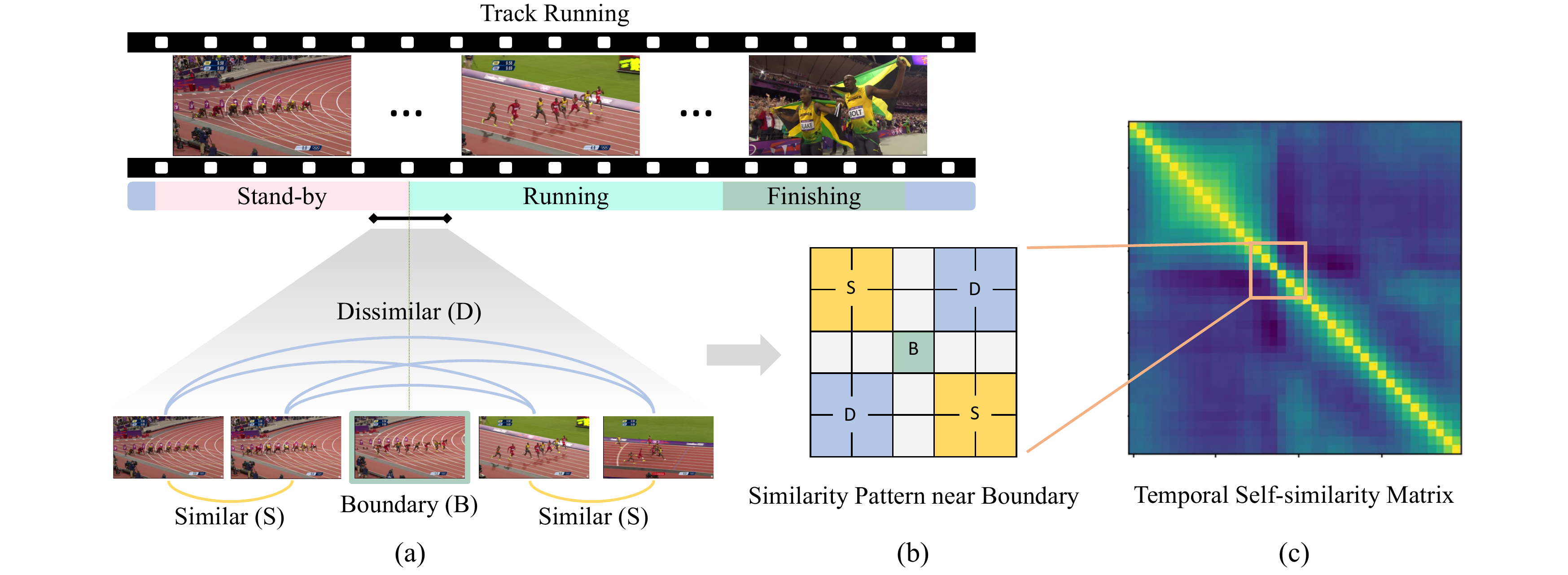}
  \caption{
  GEBD is a task of finding boundaries of one level deeper granularity compared to the video level event.
  (a) The relationship between frames showing that the \emph{local} similarities between adjacent frames are maintained except near the event boundary. 
  (b) Similarity pattern near a event boundary \textit{B} where the yellow region \textit{S} indicates high similarity score and the blue region \textit{D} indicates low similarity score. 
  (c) Temporal Self-similarity Matrix (TSM) represents the pairwise self-similarity scores between video frames.
  Similar patterns are observed in event boundaries, and we can detect boundary frames by mining this pattern from the TSM. 
  }
  \label{fig:teaser}
\end{figure*}

\section{Introduction}
\label{sec:intro}
With the proliferation of video platforms, video understanding tasks are drawing substantial attention in computer vision community.
The prevailing convention for video processing~\cite{buch2017sst, gao2017turn,Lin_2018_ECCV, escorcia2016daps, kang2021cag} is still dividing the whole video into short non-overlapping snippets with a fixed duration, which neglects the semantic continuity of the video.
On the other hand, cognitive scientists have observed that human senses the visual stream as a set of events~\cite{tversky2013event}, which alludes that there is room for research to find out a video parsing method that preserves semantic validity and interpretability of video snippets.

From this perspective, Generic Event Boundary Detection (GEBD)~\cite{shou2021gebd} can be seen as a new attempt to interconnect human perception mechanisms to video understanding.
The goal of GEBD is to pinpoint the content change moments in which humans perceive them as event boundaries.
To reflect human perception, labels for the task are annotated following the instruction of finding boundaries of  \textit{one level deeper} granularity compared to the video-level event, regardless of action classes.
This characteristic differentiates GEBD from the previous video localization tasks~\cite{xia2020survey} in the sense that the labels are only given by natural human perception, not the predefined action classes.

Recently released Kinetics-GEBD~\cite{shou2021gebd} is the first dataset for the GEBD.
The distinctive point in the dataset is that the event boundary labels are annotated by 5 different annotators, making the dataset convey the subjectivity of human perception.
Besides, various baseline methods for the GEBD task were also included in~\cite{shou2021gebd}.
Based on the similarity to Temporal Action Localization (TAL), many of the suggested GEBD methods were extensions of previous TAL works~\cite{lin2019bmn, lea2016segmental}, while several unsupervised methods exploited shot detection methods and event segmentation theory~\cite{kurby2008segmentation,reynolds2007computational,zacks2007event}.
Detailed explanations about baseline GEBD methods can be found in \Sref{sec:related_work}.

Existing methods have a clear limitation in that they directly use pretrained features to predict boundary points.
The features are extracted from the classification-pretrained networks like ResNet-50~\cite{he2016deep}, so that they inevitably contain class-specific or object-centric information.
For capturing generic event boundaries, such aspects of the features can act like noise, not providing meaningful cues for finding event boundaries.



In this paper, we introduce a novel method to discover generic event boundaries, which can be put into both \textit{unsupervised} and \textit{supervised} settings.
Our main intuition comes from observing the self-similarity of a video, visualized as Temporal Self-similarity Matrix (TSM).
While TSM has been considered as a useful tool to analyze \emph{periodic} videos due to its robustness against noise~\cite{dwibedi2020counting, panagiotakis2018unsupervised, benabdelkader2001eigengait, benabdelkader2004gait}, we found that TSM's potential is not limited to periodic videos, but can also be extended to analyzing \emph{non-periodic} videos if we focus on its local diagonal patterns.
To be specific, we can exploit TSM's robustness by taking it as an information bottleneck for the GEBD solver, making it perform well on unseen scenes, objects, and even action classes~\cite{dwibedi2020counting}.


Figure \ref{fig:teaser} briefly illustrates our observation.
For a sequence of events in a given video, there is a semantic inconsistency at the event boundary, resulting in the similarity-break near the boundary point.
As TSM depicts self-similarity scores between video frames, this similarity-break brings distinctive patterns (\Fref{fig:teaser} (c)) on TSM, which can be a meaningful cue when it comes to detecting event boundaries. 
Hence, we exploit TSM as the \emph{final} representation of the given video and devise a novel method to detect event boundaries, namely Recursive TSM Parsing (RTP).
Paired with our Boundary Contrastive loss (BoCo loss) , RTP can be extended to \emph{Unsupervised} Boundary Contrastive (UBoCo) learning, a fully label-free training framework for event boundary detection.

Going one step further, we also propose \emph{Supervised} Boundary Contrastive (\textbf{SBoCo}) learning approach for the GEBD task, which utilizes TSM as an interpretable \textit{intermediate} representation.
Unlike UBoCo that uses an algorithmic method to parse TSM, this supervised approach has TSM decoder, which is a standard neural network.
By merging binary cross-entropy (BCE) and BoCo loss, our supervised approach achieved the state-of-the-art performance in recent official GEBD challenge\footnote{CVPR'21 LOng-form VidEo Understanding (LOVEU) Kinetics-GEBD challenge}.

To summarize, the main contribution of the paper is as follows:
\begin{itemize}
     \item We discovered that the properties of Temporal Self-Similarity Matrix (TSM) matches very well with the Generic Event Boundary Detection (GEBD) task, and propose to use TSM as the representation of the video to solve GEBD.  
     \item Taking advantage of TSM's distinctive boundary patterns, we propose Recursive TSM Parsing (RTP) algorithm, which is a divide-and-conquer approach for detecting event boundaries. 
     \item Unsupervised framework for GEBD is introduced by combining RTP and the new Boundary Contrastive (BoCo) loss. Using the BoCo loss, the video encoder can be trained without labels and generate more distinctive TSMs. Our unsupervised framework outperforms not only previous unsupervised methods, but also supervised methods. 
     \item Our framework can be easily extended to the supervised setting by adding a decoder and achieve the state-of-the-art performance by a large margin (16.2\%).
\end{itemize}

\section{Related Work}
\label{sec:related_work}
\subsection{Generic Event Boundary Detection}
Generic Event Boundary Detection (GEBD)~\cite{shou2021gebd} is a newly introduced video understanding task that aims to spot event boundaries that coincide with human perception.
GEBD shares an important trait with the popular video event detection task, Temporal Action Localization (TAL) in that the model should be aware of the action happening in the video.
Among many TAL methods, BMN\cite{lin2019bmn} has the intermediate stage of calculating the probability of the start and the end point of an action instance, making its extension to GEBD more feasible.
Therefore,~\cite{shou2021gebd} introduced a method called BMN-StartEnd, which treated BMN model's intermediate action start-end detection results as event boundaries.

Apart from extending TAL, better performance can be achieved by treating the GEBD task as a framewise binary classification (boundary or not).
For the network architecture, temporal modeling using TCN~\cite{lea2016segmental, lin2018bsn} has been tested, but a simple linear classifier with concatenated average feature (denoted as PC in~\cite{shou2021gebd}) resulted in the best performance.

For unsupervised GEBD, a straightforward method is to utilize a previous shot boundary detector~\footnote{https://github.com/Breakthrough/PySceneDetect}. 
However, generic event boundaries consist of various kind of event boundaries including change of action,  subject, and environment, implying that only a small portion of event boundaries can be detected with the shot boundary approach.
Thus,~\cite{shou2021gebd} devised a novel unsupervised GEBD solver exploiting the fact that predictability is the main factor in human's event perception~\cite{kurby2008segmentation}.
Among many suggested unsupervised methods, the PA (PredictAbility) method resulted in the best performance.

Except for the PA method, many of the suggested GEBD approaches are just straightforward extensions of previous methods that targeted other video tasks, raising the need for a  GEBD-specialized solution.
Focusing on the GEBD's distinctive characteristics, we suggest a novel method that exploits TSM representation, which shows a unique pattern near the boundary.

\subsection{Temporal Self-similarity Matrix}
With the increasing popularity of using self-attention for its benefits~\cite{vaswani2017attention,devlin2018bert,dosovitskiy2020image,carion2020end}, Temporal Self-similarity Matrix (TSM) has also been getting much attention lately as an interpretable intermediate representation for video. 
From a given video with $L$ frames, each value at position ($i$, $j$) in TSM is calculated using cosine or L2-distance similarity between frame $i$ and $j$, resulting in an $L \times L$ matrix.
As TSM represents the similarity between different frames in a video, it is often used for the task of repetition counting~\cite{dwibedi2020counting, panagiotakis2018unsupervised}, gait recognition~\cite{benabdelkader2001eigengait, benabdelkader2004gait}, and language-video localization~\cite{nam2021zero}.
Furthermore, as TSM effectively reflects temporal relationships between features, some works that require a general video representation like action classification~\cite{chen2020hierarchical}, and representation learning~\cite{kordopatis2019visil} also utilize TSM as the intermediate representation.
As the local temporal relationship is the key feature for the event boundary detection, our method also utilizes TSM as the video representation, enhancing the overall performance compared to previous methods.

\subsection{Contrastive representation learning}
Contrastive learning, with its generality and simplicity, is gaining increasing attention in computer vision society. 
Its main idea is to attract semantically matching samples closer and repel mismatching ones.
Note that the method of determining whether a pair is semantically matching or not is not fixed in contrastive learning.
Thus, while many recent works~\cite{he2020momentum, chen2020improved, chen2021exploring, grill2020bootstrap, chen2020big} combined contrastive learning with data augmentation and treated it as a self-supervised approach, some works extended contrastive learning to standard supervised learning~\cite{khosla2020supervised, cui2021parametric}.
As our boundary contrastive loss utilizes (pseudo-) boundary labels for contrastive learning, it has a close relationship with the supervised contrastive learning.

Furthermore, there are various works on contrastive video representation learning, which viewed contrastive learning from temporal~\cite{pan2021videomoco}, spatio-temporal~\cite{qian2021spatiotemporal}, and temporal-equivariant~\cite{jenni2021time} perspectives. 
Especially, ~\cite{chen2021shot} applied self-supervised contrastive pretext task to learn appropriate feature representation, proving its effectiveness on detecting shot frames.
However, it still needs \emph{downstream} task training (supervised learning) to get final results, while our approach directly yields event boundaries.

\begin{figure*}
  \includegraphics[width=\textwidth]{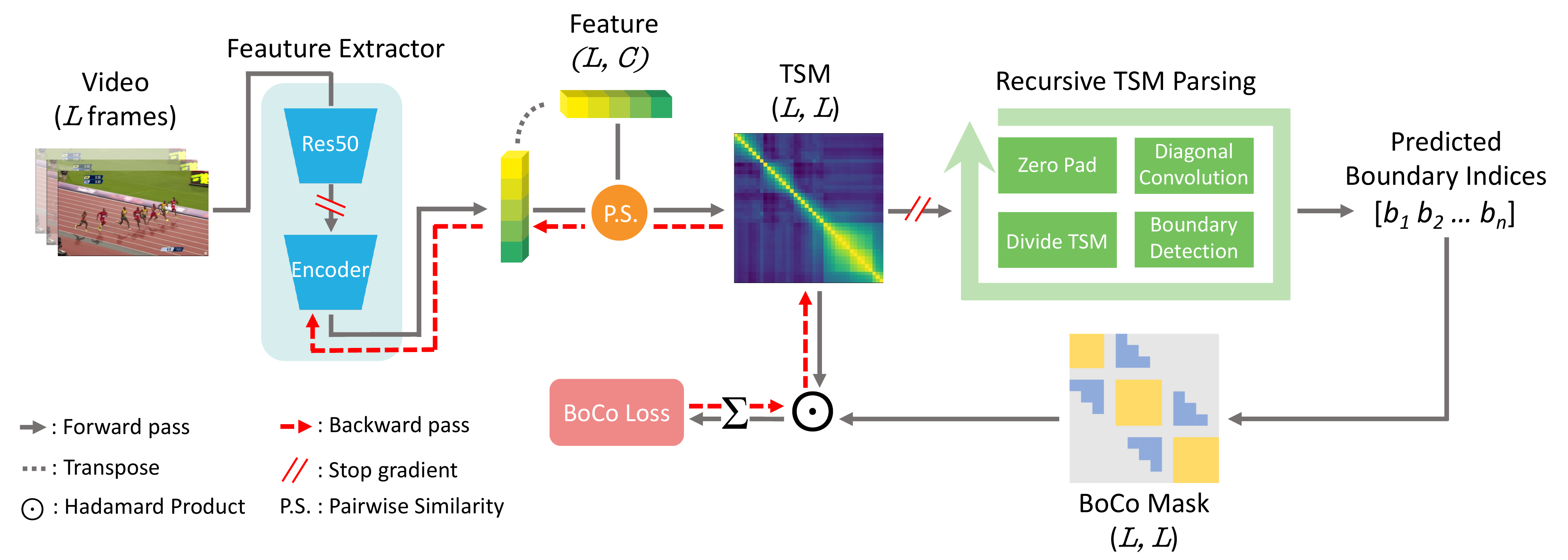}
  \caption{Overview of the Unsupervised Boundary Contrastive (UBoCo) learning framework for GEBD.}
  \label{fig:overview}
\end{figure*}

\section{Proposed Method}
\label{sec:proposed_method}
Given a video with $L$ frames, GEBD solver returns a list containing event boundary frame indices. 
In this section, we introduce our novel unsupervised/supervised GEBD solver, which makes use of TSM in the intermediate stage.

\begin{figure}
    \centering
    \includegraphics[width=\linewidth]{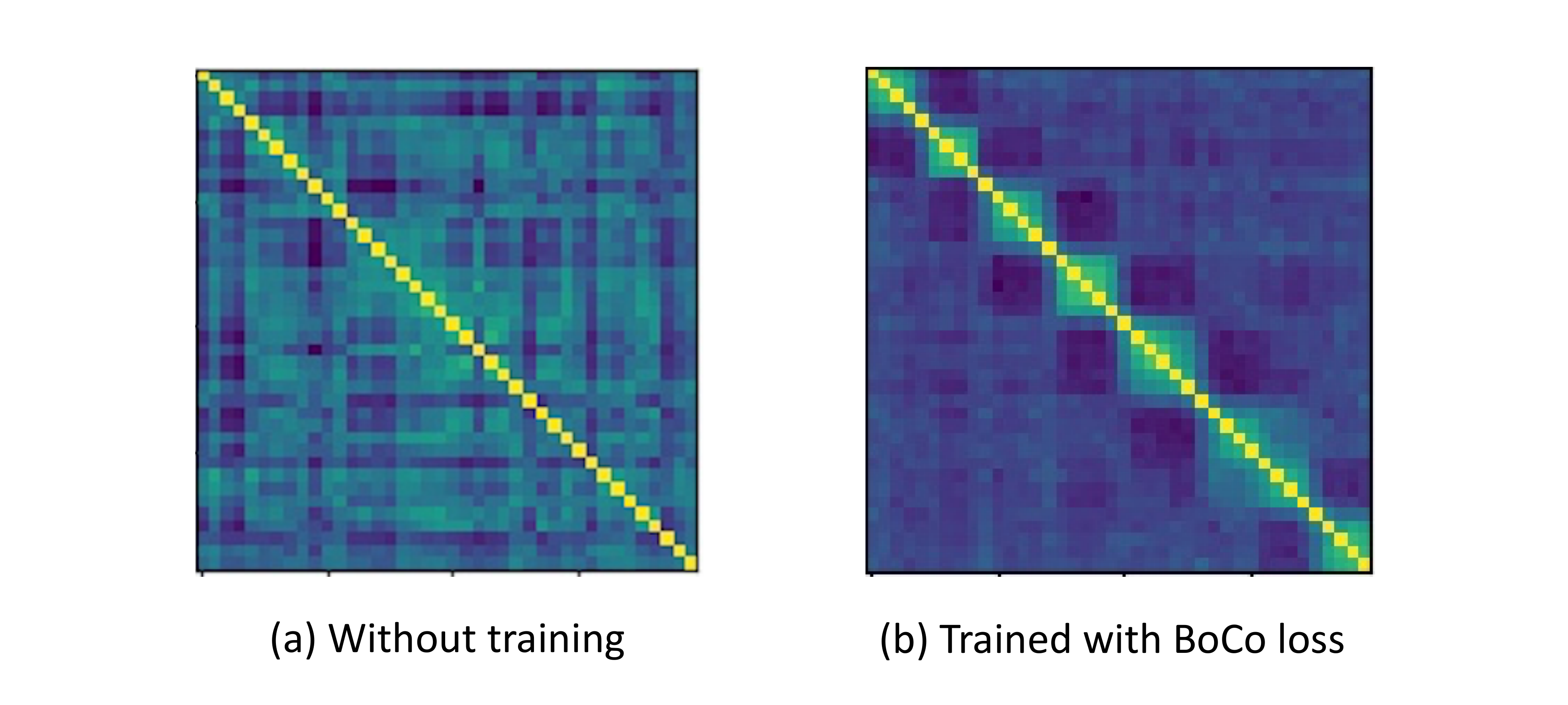}
    \caption{
    (a) Noisy TSM can be observed at the early stage of training. (b) As training proceeds, the TSM is getting sharper, showing distinctive boundary patterns.
    }
    \label{fig:tsm}
\end{figure}

\subsection{Overview}
\label{subsec:UBoCo_overview}
\Fref{fig:overview} shows an overview of our Unsupervised Boundary Contrastive learning framework (UBoCo).
In the first stage of the procedure, framewise features are extracted from the raw video frames, using a neural feature extractor.
The feature extractor consists of a pretrained frame encoder (ImageNet pretrained ResNet50~\cite{he2016deep}) and an extra encoder.
Weights are fixed for the pretrained encoder, while the custom encoder's weights are trainable.

With the extracted feature, self-similarities between the frames are computed, forming Temporal Self-similarity Matrix (TSM).
Given a TSM, Recursive TSM Parsing (RTP) algorithm (\Sref{subsubsec:RTP}) produces boundary indices prediction.
Considering this prediction as a pseudo-label, the encoder can be trained with the standard gradient-descent algorithm using the BoCo loss (\Sref{subsubsec:boundary_contrastive_loss}), which enriches the encoder to produce boundary-sensitive features.
Note that gradients directly flow to the TSM, bypassing the non-differentiable RTP pass.

Our concept of pseudo-label framework resembles the popular k-means clustering algorithm~\cite{macqueen1967some}.
In k-means algorithm, the mean values of clustered vectors in the previous stage become the new criterion for the current clustering stage.
The centroids of k-means clustering are analogous to pseudo-labels of our method in the sense that the current training step is conducted based on the result of the previous step.

In the early stage of training, TSM is not perfectly discriminative due to undertrained feature encoder (\Fref{fig:tsm}).
At this stage, only obvious boundaries that involve drastic visual changes can be detected by the RTP algorithm. 
With more training, the BoCO loss enables the feature encoder to produce more robust TSM (\Fref{fig:tsm}), making the discriminative power of TSM stronger. 
As a better feature encoder generates better quality pseudo-labels, the BoCo loss also becomes more powerful as the training proceeds. 
This progressive improvement is the key property of our UBoCo framework, and its effect on the overall performance will be shown in the Experiments Section. 

\begin{figure*}
  \includegraphics[width=\textwidth]{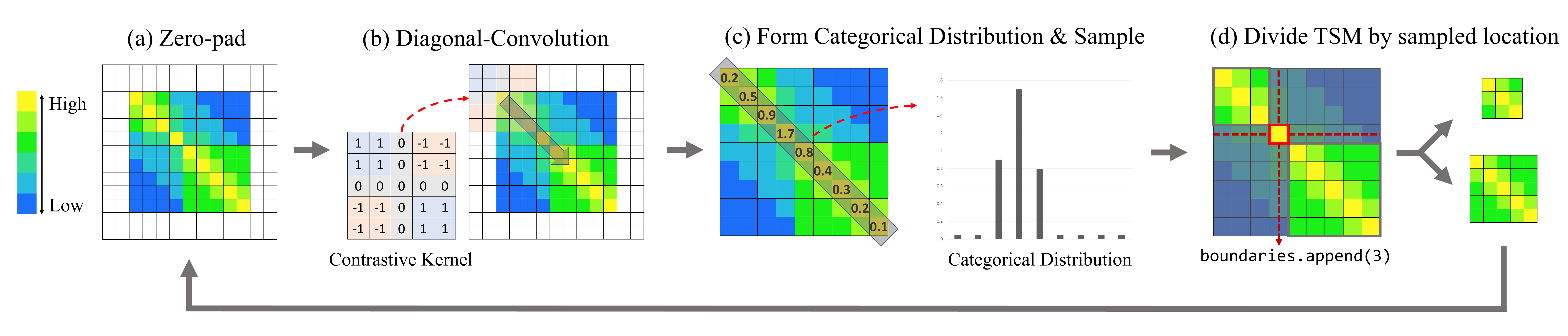}
  \caption{The figure demonstrates how Recursive TSM Parsing (RTP) works, assuming a 9x9 TSM and a 5x5 contrastive kernel. The yellow region in TSM indicates a high similarity score, while the blue region means a lower score. Using the output boundary index, TSM is divided into smaller TSMs, which again go through the same process. Note that the recurrence stops for the small sub-TSM in this example as it reached a predefined minimal length. 
  }
  \label{fig:RTP}
\end{figure*}

\subsection{Unsupervised Boundary Contrastive Learning}
\label{subsec:unsupervised_boundary_contrastive_learning}
\subsubsection{Recursive TSM Parsing}
\label{subsubsec:RTP}
With the intuition illustrated in Figure~\ref{fig:teaser}, we devised a novel method called Recursive TSM Parsing (RTP), to detect event boundaries from a given TSM.
As a divide-and-conquer approach, RTP algorithm takes TSM as its input and yields boundary indices in a \emph{recursive} manner as shown in \Fref{fig:RTP}.

First, the input TSM is zero-padded to make convolution operation to be applicable on corner elements (\Fref{fig:RTP} (a)).
Then, diagonal elements of TSM are convolved with the ``contrastive kernel'', which is a materialization of the boundary pattern in~\Fref{fig:teaser}.
After the diagonal convolution with the contrastive kernel, scalar values that represent boundariness are produced (Figure~\ref{fig:RTP} (b)).
A higher boundary score means that the local TSM pattern matches well with the contrastive kernel, indicating a higher probability of being an event boundary.
Once boundary scores are computed, all but the scores affected by the zero-padding are shared through multiple RTP passes.

With the computed boundary scores, we then select an index that corresponds to an event boundary.
For this process, we form a categorical distribution of the boundary scores.
To reduce excessive randomness, we only preserve top k\% scores.
The boundary frame index is then determined by sampling with the computed distribution  (\Fref{fig:RTP} (d)). 
By substituting the straightforward max operation with this sampling strategy, we can diversify training samples, compensating for the training data limitation.
Now with the boundary frame index given, the TSM is \emph{divided} into two separate TSMs, each of which is forwarded to another run of the RTP algorithm  (\Fref{fig:RTP} (d)).

The above pass is recursively executed until one of the following end-conditions is satisfied: $a$) the parsed TSM is smaller than a predefined threshold $T_1$, or $b$) difference between the max boundary score and the mean boundary score is smaller than a threshold $T_2$.
The first condition represents a prior assumption on the minimal length of an event segment, and the second condition handles long event segment cases.
Note that a small difference between the highest score and the mean value implies that there are no distinguishable points.

Although it may be seen as a minor step, zero-padding plays an important role in RTP.
Corner elements of the TSM are one of the following: start, end point of the video, or proximate point to the detected boundary.
This indicates that corner points are unlikely to be a boundary frame.
Therefore, in addition to enabling the boundary computation at corners, the zero-padding also allows for assigning relatively lower boundary scores for the corners, suppressing false or duplicated event boundary detection.

%

\subsubsection{Boundary Contrastive Loss}
\label{subsubsec:boundary_contrastive_loss}
Similar to RTP,  Boundary Contrastive loss (BoCo loss) shares the same ``boundary pattern'' intuition in Figure~\ref{fig:teaser}.
The objective of BoCo loss is to train the feature encoder, in order to produce informative TSM, whereas the goal of RTP is to extract boundary indices from a given TSM.
With the boundary indices computed by RTP, the BoCo loss helps TSM to be more distinguishable at boundaries (\Fref{fig:tsm}).

\begin{figure}
    \centering
    \includegraphics[width=\linewidth]{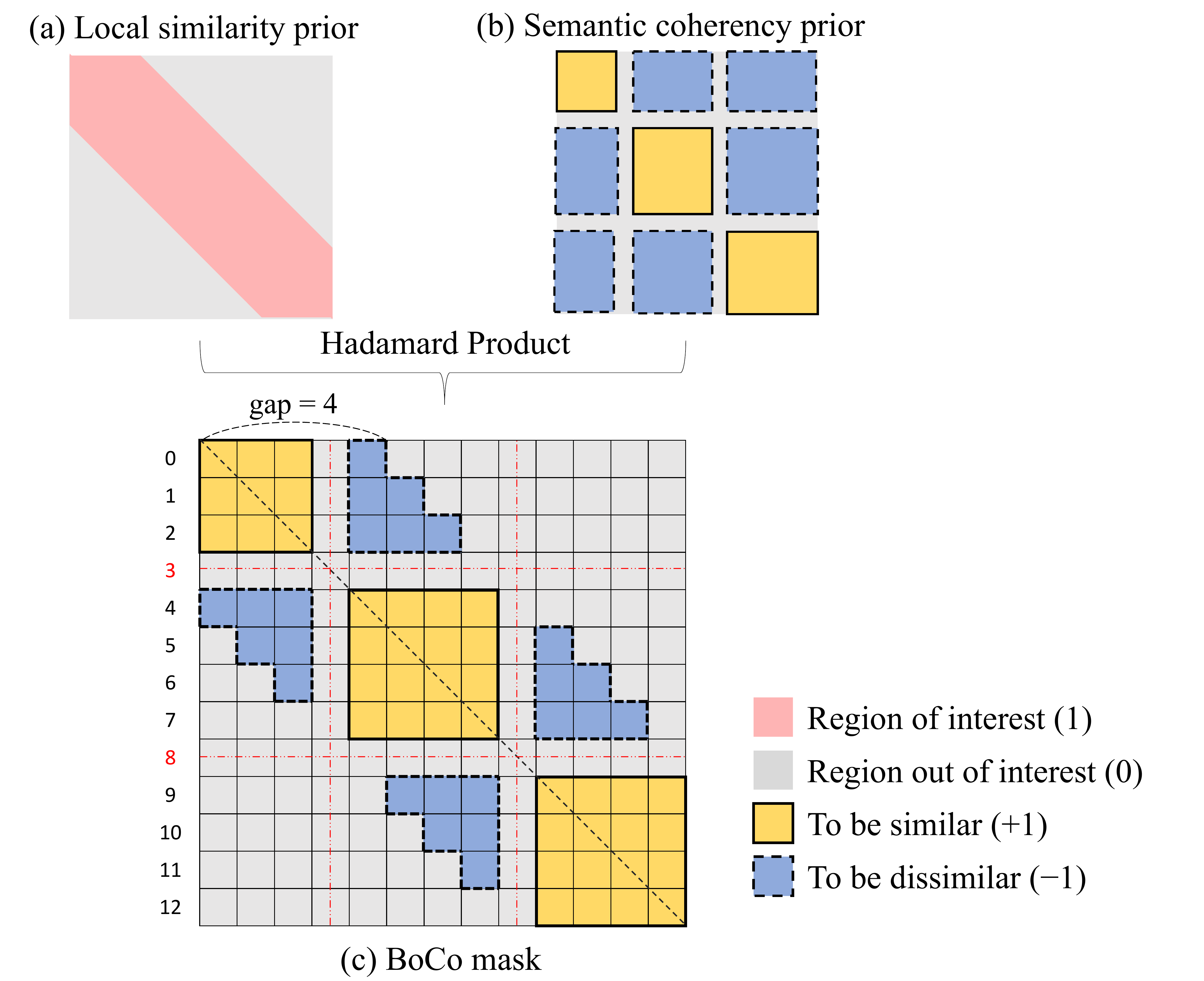}
    \caption{
    Given the list of boundary indices ([3,8] in the above example), (c) represents the mask to compute BoCo loss. Both prior (a), (b) are merged by element-wise multiplication.
    }
    \label{fig:boco_loss}
\end{figure}

Inspired by the recent success of contrastive learning, BoCo loss adopts the metric learning strategy.
\Fref{fig:boco_loss} explains how positive and negative samples are selected for the BoCo loss.
As our task focuses on the short-term frame relationship, the similarity between distant frames would not give much information for detecting event boundaries.
From this assumption, we pose \textit{local similarity prior} as described in \Fref{fig:boco_loss} (a).
The prior implies that we only care about the similarities between frames within a predefined interval, or in other words, ``gap''.

With given $n$ boundary frame indices, a video can be parsed into $n+1$ segments.
Recall that the video frames in the same segment are \emph{semantically coherent}, meaning that the frame similarity among them should be high, while the similarity between the frames in different segments should stay low.
This assumption can be implemented as the \textit{semantic coherency prior} mask shown in  \Fref{fig:boco_loss} (b).

By using elementwise multiplication, we can combine both assumptions, getting the BoCo mask in \Fref{fig:boco_loss} (c) that represents valid positive/negative pairs.
Once the positive/negative pairs are determined, there are several choices for the metric learning, including ~\cite{khosla2020supervised}.
However, for computational efficiency and straightforward implementation, BoCo loss is simply computed as the difference between the mean values of the blue region and the yellow region in \Fref{fig:boco_loss} (c).
Even though this simple approach gives satisfactory results (Section~\ref{sec:experiments}), the metric learning loss function for our algorithm could be a direction for our future work.

\subsection{Supervised Boundary Contrastive Learning with Decoder}
\label{subsec:boundary_contrastive_learning_with_decoder}
\begin{figure}
    \centering
    \includegraphics[width=\linewidth]{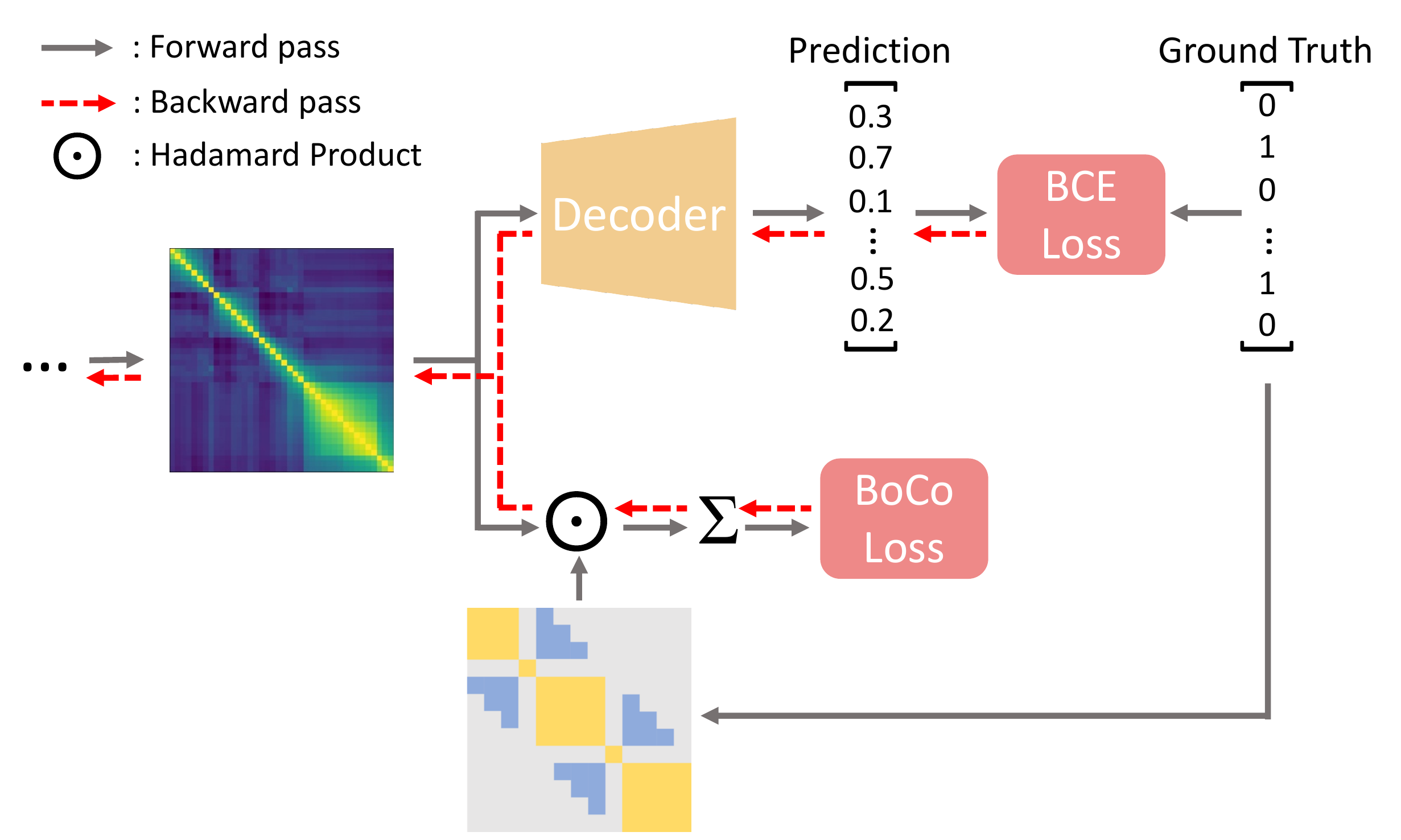}
    \caption{Supervised GEBD framework with a decoder. With supervision, we can replace RTP with a neural decoder and improve the GEBD performance with additional BCE loss.}
    \label{fig:SBoCo}
\end{figure}

By simply replacing the pseudo-label with the ground-truth label, UBoCo can be converted into Supervised Boundary Contrastive learning (SBoCo).
While this straightforward version of SBoCo works quite well, we can substitute RTP algorithm with TSM decoder (\Fref{fig:SBoCo}), which consists of convolutional neural networks~\cite{he2016deep} and transformers~\cite{vaswani2017attention}.
In this approach, probabilities of being event boundary can be directly acquired from the decoder, only requiring simple post-processing (e.g. thresholding).

There are several advantages to adopting a neural TSM decoder.
First, another widely used loss term, binary cross-entropy loss, can be additionally utilized.
It enables the model to receive more informative gradient signals during the training procedure.
Moreover, as it allows direct prediction, we can detour the recursive procedure of RTP, making faster inference possible.
Lastly, thanks to the strong representation power of neural networks, we can employ multi-channel TSM, compensating for the limitation of the single-channel TSM's expressive power.
This approach shares the fundamental idea with~\cite{dwibedi2020counting}, which used TSM as an interpretable intermediate representation.
More details on the decoder will be provided in the supplementary materials. 

\section{Experiments}
\label{sec:experiments}

\subsection{Benchmark Dataset} 
\label{subsec:benchmark_dataset}
Kinetics-GEBD consists of train, validation, and test sets, each with about 18K videos from Kinetics-400 dataset~\cite{kay2017kinetics}.
Since the ground truth for the test set is not released, we use the validation set of Kinetics-GEBD as the test set as in ~\cite{shou2021gebd} and split the train set of Kinetics-GEBD into the train (80\%) and the validation (20\%) for our experiments. 
As for the evaluation metric, we validate our experiments on  F1@0.05.
The value of 0.05 is the threshold of the Relation Distance (Rel.Dis.).
Given a Rel.Dis., the predicted point is treated as correct if the discrepancy between the predicted and the ground truth timestamps is less than the threshold.
Following the benchmark~\cite{shou2021gebd} and the official challenge, we analyze ours results mainly on this threshold value.
In \Tref{tab:results_on_gebd}, we also provide the average F1 score, which is the mean of the F1 scores using the thresholds from 0.05 to 0.5 with 0.05 interval.
As we show in the supplementary materials, the results for other thresholds still show the competitiveness of our methods.

\begin{table}[]
    \centering
    \resizebox{\linewidth}{!}{%
    \begin{tabular}{cccc}
    \toprule
         & Method & F1@0.05 & Average F1\\
         \hline
        \multicolumn{1}{c|}{\multirow{5}{*}{Unsupervised}} & SceneDetect & 27.5 & 31.9\\
        \multicolumn{1}{c|}{} & PA-Random & 33.6 & 50.6\\
        \multicolumn{1}{c|}{} & PA & 39.6 & 52.7 \\ 
        \multicolumn{1}{c|}{} & UBoCo-Res50 (ours) & \textbf{70.3} & \textbf{86.7}\\
        \multicolumn{1}{c|}{} & UBoCo-TSN (ours) & \textbf{70.2} & \textbf{86.7}\\
        \hline
        \multicolumn{1}{c|}{\multirow{7}{*}{Supervised}} & BMN & 18.6 & 22.3\\
        \multicolumn{1}{c|}{} & BMN-StartEnd & 49.1 & 64.0\\
        \multicolumn{1}{c|}{} & TCN-TAPOS & 46.4 & 62.7\\ 
        \multicolumn{1}{c|}{} & TCN & 58.8 & 68.5\\
        \multicolumn{1}{c|}{} & PC & 62.5 & 81.7\\
        \multicolumn{1}{c|}{} & SBoCo-Res50 (ours) & \textbf{73.2} & \textbf{86.6}\\
        \multicolumn{1}{c|}{} & SBoCo-TSN (ours) & \textbf{78.7} & \textbf{89.2}\\
    \bottomrule
    \end{tabular}
    }
    \caption{Results on Kinetics-GEBD for unsupervised (top) and supervised (bottom) methods. The scores of previous methods are from ~\cite{shou2021gebd}.}
    \label{tab:results_on_gebd}
\end{table}

\subsection{Implementation Details}
\label{subsec:implementation_details}
We used ResNet-50~\cite{he2016deep} pretrained on ImageNet~\cite{deng2009imagenet} and the weights from torchvision~\cite{marcel2010torchvision} as our main backbone for fair comparisons with previous methods~\cite{shou2021gebd}, which also used the same features.
We additionally tested with TSN~\cite{wang2016temporal} pretrained on Kinetics dataset~\cite{kay2017kinetics} to maximize the model performance.
All experiments were conducted on a single Nvidia RTX 2080 Ti GPU equipped machine.
We trained our models with the batch size of 32 using AdamW optimizer with the learning rate of 1e-3.
Our model's encoder consists of 1D CNNs and Mixer~\cite{tolstikhin2021mlp} for capturing short-term and long-term representations respectively.
More details about our feature extraction and models are described in the supplementary materials.

\subsection{Results on Kinetics-GEBD}
\label{subsec:reults_on_gebd}

\Tref{tab:results_on_gebd} illustrates the results of our models compared to prior works in unsupervised and supervised conditions. 
Our models outperform previous models by a large margin for not only the unsupervised setting (30.7\% higher than the SOTA) but also the supervised setting (10.7\% better than the SOTA) with the same extracted features (ResNet-50). 
We can demonstrate that our \textit{unsupervised} model, UBoCo, is extremely powerful to surpass the previous \textit{supervised} state-of-the-art model by a good margin of 7.8\%.
We notice that with the TSN features, extracted by the video-level model TSN, there is a remarkable improvement for the supervised setting, increasing the performance by 16.2\% over the previous SOTA. 

\begin{figure}
    \centering
    \includegraphics[width=0.9\linewidth]{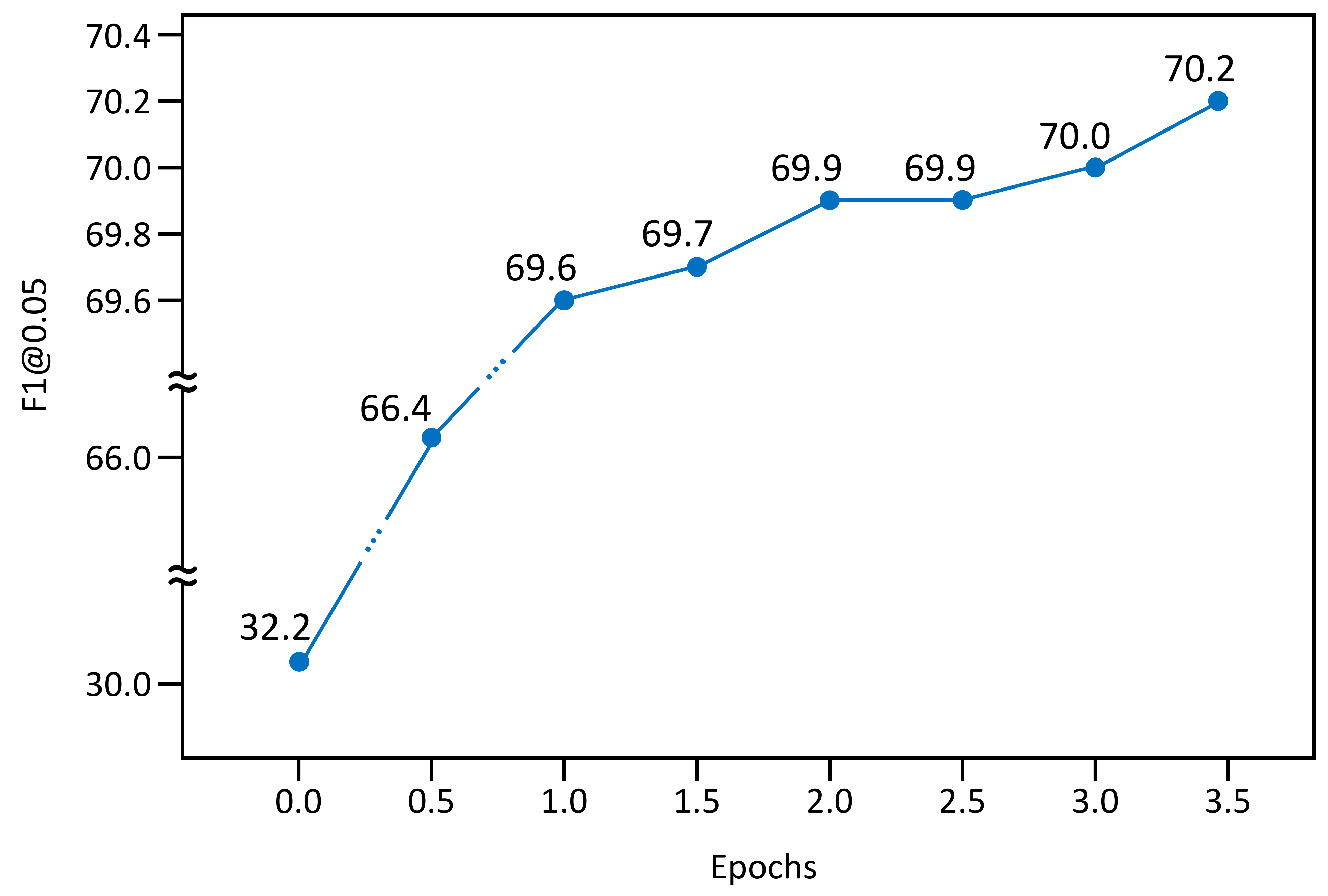}
    \caption{Improvement of UBoCo as the encoder of the model is trained on pseudo-labels in a self-supervised manner. } 
    \label{fig:ablation-ssl}
\end{figure}

\subsection{Ablation Studies}
\label{subsec:ablation_studies}

\subsubsection{Self-supervision with Pseudo-label}
\label{subsubsec:ablation_self_supervision}
From the unsupervised model (UBoCo), we can make the pseudo-labels as explained in \Sref{subsec:unsupervised_boundary_contrastive_learning}.
Using these pseudo-labels, we can train our UBoCo model with gradient-descent, which can be seen as a self-supervised approach.
The progressive improvement of the performance is illustrated in \Fref{fig:ablation-ssl}.
We can observe that the accuracy drastically increases as the training evolves, showing the power of self-supervision with BoCo loss.
Another interesting observation is the performance at epoch 0 (33.2\%), which is comparable to the existing state-of-the-art unsupervised model (39.6\%).
It indicates that even with undertrained feature encoder, our RTP can catch some obvious event boundaries.



\begin{table}[]
\centering
\begin{tabular}{c|cc}
\toprule
 & UBoCo-Res50 & UBoCo-TSN \\ \hline
Thresholding & 27.3 & 27.3 \\
Local Maxima & 68.1 & 68.8 \\
RTP w.o. ZP & 55.3 & 54.8 \\
(ours) RTP & 70.3 & 70.2 \\ 
\bottomrule
\end{tabular}
\caption{F1@0.05 scores corresponding to different parsing algorithms. ZP stands for Zero-Padding. They share the same procedure until the boundary scores are calculated. For the method \textit{Thresholding}, the points whose sigmoid value of event score is larger than the threshold are predicted as event boundaries. We use the best score for various thresholds from 0.1 to 0.5.}
\label{tab:RTP}
\end{table}

\begin{table}[]
    \centering
    \resizebox{0.9\linewidth}{!}{%
    \begin{tabular}{cc|cc}
    \toprule
        Supervision & Decoder & SBoCo-Res50 & SBoCo-TSN \\
        \hline
        & & 70.3 & 70.2\\
        \checkmark & & 71.1 (+0.8) & 75.5 (+5.3)\\
        \checkmark & \checkmark & \textbf{73.2 (+2.9)} & \textbf{78.7 (+8.5)}\\
    \bottomrule
    \end{tabular}
    }
    \caption{F1@0.05 scores for the effect of supervision and decoder.}
    \label{tab:supervision_and_decoder_layer}
\end{table}

\begin{figure*}
    \centering
    \includegraphics[width=\textwidth]{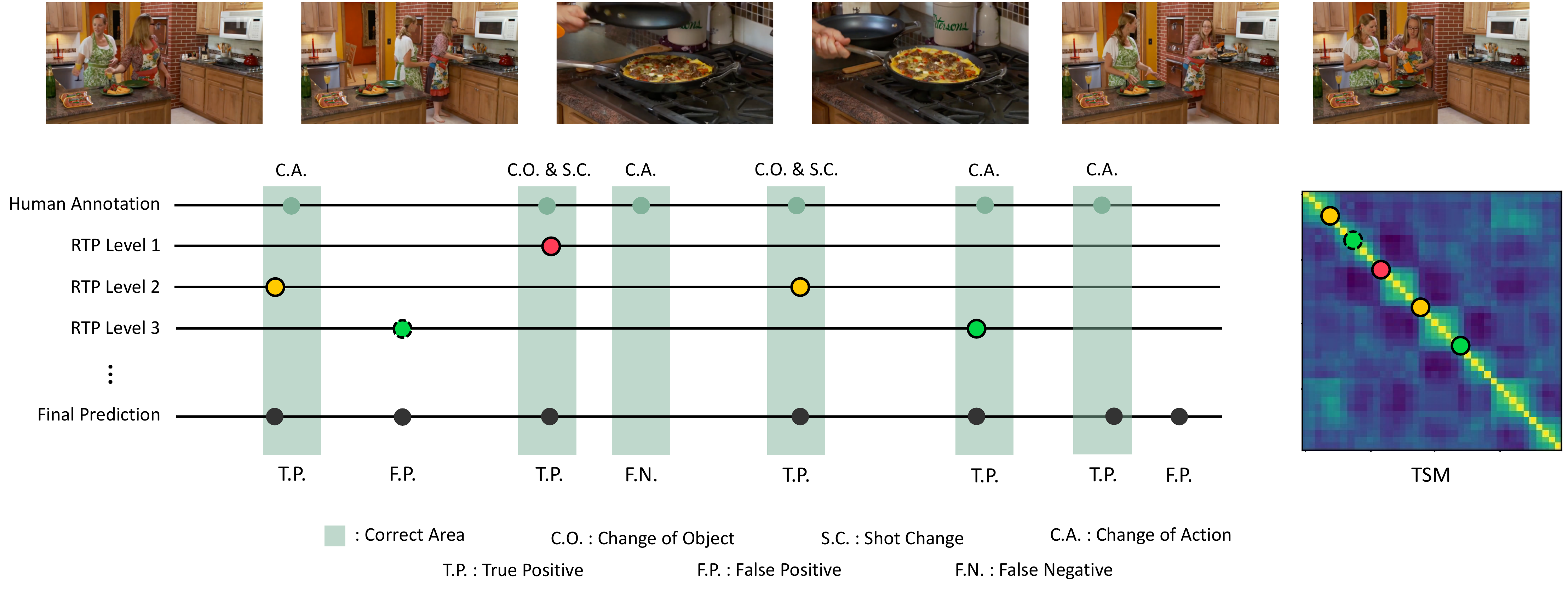}
    \caption{Above figure illustrates how RTP detects event boundaries from the given TSM. As shown in the figure, apparent boundaries including shot change are captured at the early level of RTP, while more subtle boundaries are deferred to the last level.}
    \label{fig:RTP_qualitative}
\end{figure*}

\subsubsection{Recursive TSM Parsing}
\label{subsubsec:RTP_exp}
The effectiveness of the Recursive TSM Parsing (RTP) algorithm with zero-padding against other straight-forward algorithms is shown in \Tref{tab:RTP}.
To extract event boundaries from a TSM, some other options include simple thresholding and finding local maxima of diagonal boundary scores ~\cite{shou2021gebd}.
As can be observed, RTP yields the best performance and zero-padding is essential for RTP process. We attribute the performance improvement to its coincidence with the GEBD's underlying assumption, ``one level deeper semantics''.
In other words, the recursive fashion of RTP easily catches the inherent hierarchy of action segments in that each recursion stage represents a different level of hierarchy.
\Fref{fig:RTP_qualitative} qualitatively illustrates the performance of RTP. 
It shows that RTP detects big changes in the early stages and starts to pick up subtle changes in an iterative way.


\subsubsection{Extension to Supervised Model (SBoCo)}
\label{subsubsec:ablation_SBoCo}
By converting the pseudo-label in \Sref{subsec:unsupervised_boundary_contrastive_learning} into human-annotated ground truth, we can outstretch the unsupervised model (UBoCo) to the supervised model (SBoCo).
Furthermore, as explained in \Sref{subsec:boundary_contrastive_learning_with_decoder}, we subjoin the decoder layer so that the model can exploit the TSM not only explicitly but also implicitly.
The results for both methods are shown in \Tref{tab:supervision_and_decoder_layer}, demonstrating that the supervision and the decoder layer help the performance enhancement, especially for the SBoCo-TSN model.
By converting input from image-level features (ResNet-50 pretrained on ImageNet) to video-level (TSN pretrained on Kinetics), we also observed additional improvement of the performance, achieving the state-of-the-art in Kinetics-GEBD.

\begin{figure}
    \centering
    \includegraphics[width=\linewidth]{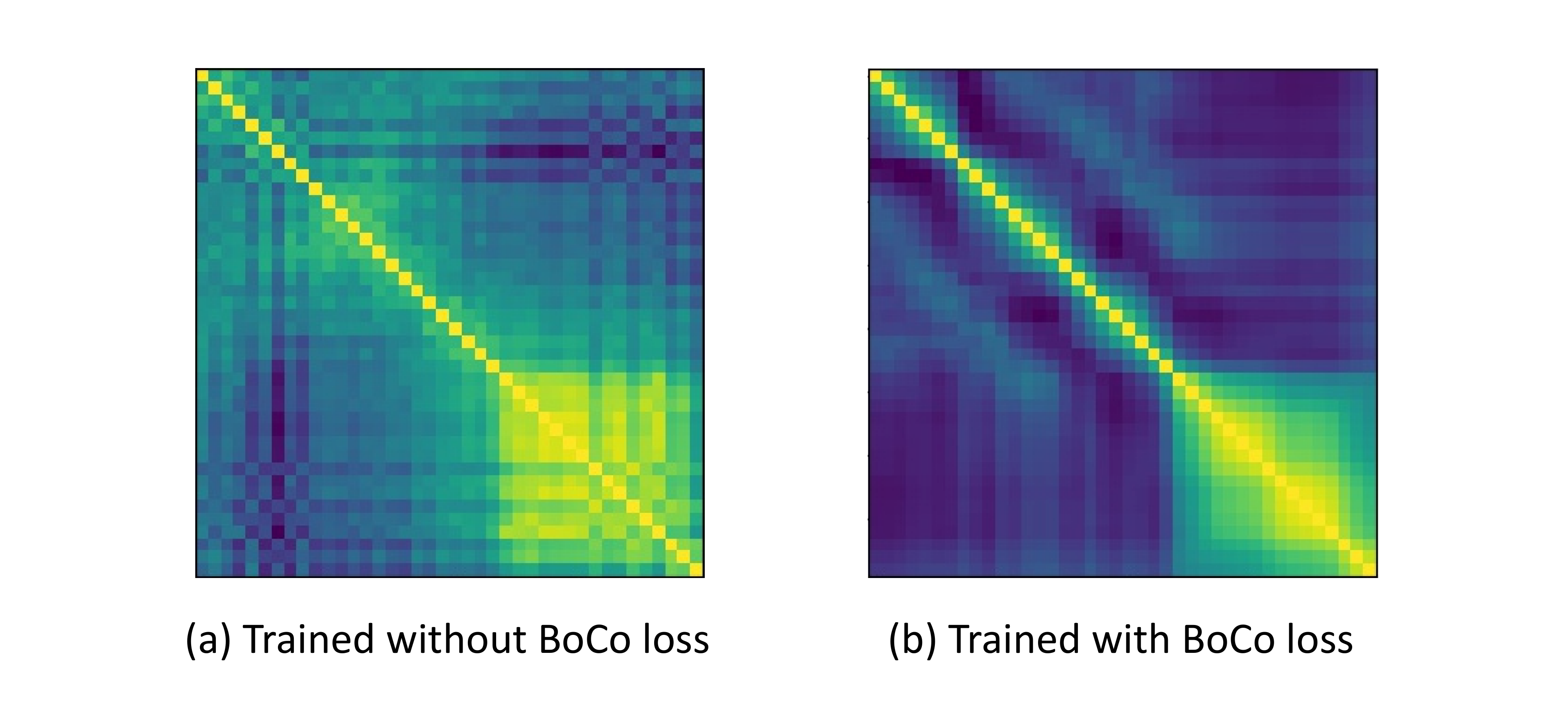}
    \caption{BoCo loss in supervised model makes the TSM more interpretable and informative. Boundary patterns in (b) is much more distinguishable than those in (a).}
    \label{fig:tsm_exp}
\end{figure}

Moreover, to validate the role of BoCo loss in the supervised setting, ablation study about BoCo loss was conducted.
Experimental result shows that accompanying BoCo loss with BCE loss can enhance the performance in both ResNet feature and TSN feature (\Tref{tab:boco_loss}).
Qualitative result (\Fref{fig:tsm_exp}) also illustrates that BoCo loss makes the TSM of a given video more interpretable.

\begin{table}[]
\centering
\resizebox{0.9\linewidth}{!}{%
\begin{tabular}{c|cc}
\toprule
 & SBoCo-Res50 & SBoCo-TSN \\ \hline
BCE loss only & 71.8 & 77.5 \\
BCE loss \& BoCo loss & \textbf{73.2 (+1.4)} & \textbf{78.7 (+1.2)} \\ 
\bottomrule
\end{tabular}
}
\caption{F1@0.05 scores without and with BoCo loss in supervised model (SBoCo).}
\label{tab:boco_loss}
\end{table}

\section{Discussion and Conclusion}
\label{sec:conclusion}
Generic Event Boundary Detection (GEBD) has the potential of being a fundamental upstream task for further video understanding in that it can make changes on prevailing video-division convention to a more human interpretable way.
Rethinking the essence of the task, we found that what actually matters for event boundaries is \emph{local} frame relationship, implying that TSM's diagonal local pattern can be a meaningful cue for boundary detection.
Expanding the idea, we proposed the novel unsupervised/supervised GEBD solver that utilizes the TSM's local distinguishing similarity pattern that emerges near the event boundaries.
Both of our methods achieved state-of-the-art results in both unsupervised/supervised GEBD settings, which promotes further research on the task.
Moreover, focusing on the fact that our unsupervised method can produce reasonable event boundaries without any human-annotated label, it has a great potential to be stretched to other unsupervised video understanding tasks.

There are also several limitations that stimulate future works.
To begin with, we used \emph{fixed} contrastive kernel as a realization of our intuition.
Despite its exceptional performance, different variation, or even learnable kernel can be adopted for future research.
Furthermore, since the only available benchmark dataset for GEBD, Kinetics-GEBD, contains videos with relatively similar duration, experiments for videos with varying duration cannot be conducted.
This calls for more GEBD labels for various kinds of videos, ranging from short Tiktok clips to full movies, in the near future.


{\small
\bibliographystyle{ieee_fullname}
\bibliography{egbib}
}

\end{document}